\documentclass[journal]{IEEEtran}

\usepackage{epsfig}
\usepackage{graphicx}
\usepackage{amsmath}
\usepackage{amssymb}
\usepackage{array}
\usepackage{booktabs}
\usepackage{multirow}
\usepackage{color}
\usepackage{cite}
\ifCLASSINFOpdf
\else
\fi
\begin{document}
%
\title{Sequence-based Person Attribute Recognition with Joint CTC-Attention Model}
%
%
%

\author{Hao Liu,
        Jingjing Wu,
        Jianguo Jiang,
        Meibin Qi,
        and~Bo Ren
\thanks{Hao Liu and Bo Ren are with the Tencent Youtu Lab, P.R.China,
e-mail: hfut.haoliu@gmail.com, timren@tencent.com.}
\thanks{Jingjing Wu, Jianguo Jiang and Meibin Qi are with Hefei University of Technology, P.R.China,
e-mail: hfutwujingjing@mail.hfut.edu.cn, jgjiang@hfut.edu.cn, qimeibin@163.com.}
}

\maketitle

\begin{abstract}
Attribute recognition has become crucial because of its wide applications in many computer vision tasks,
such as person re-identification. Like many object recognition problems, variations in viewpoints, illumination, and recognition at far distance,
all make this task challenging. In this work, we propose a joint CTC-Attention model (JCM),
which maps attribute labels into sequences to learn the semantic relationship among attributes.
Besides, this network uses neural network to encode images into sequences,
and employs connectionist temporal classification (CTC) loss to train the network with the aim of improving the encoding performance of the network.
At the same time, it adopts the attention model to decode the sequences, which can realize aligning the sequences and better learning the semantic information from attributes.
Extensive experiments on three public datasets, i.e., Market-1501 attribute dataset, Duke attribute dataset and PETA dataset, demonstrate the effectiveness of the proposed method.
\end{abstract}

\section{Introduction}

Person re-identification (person Re-ID) aims at identifying a specific person appearing at different times and places, which becomes critical to many surveillance,
security and multimedia applications, such as on-line tracking or off-line searching. As important visual clues, human attributes, including age, gender, hair, clothing style, and many other instances,
are highly semantic and informative to describe the appearance of a person.
Human attribute recognition can benefit a variety of visual surveillance applications such as person re-identification \cite{Alpher01, Alpher07, Alpher22, Alpher24},
face verification \cite{Alpher39}, and other computer vision applications, such as fine-grained recognition \cite{Alpher38},
object categorization \cite{Alpher48}, object description \cite{Alpher49}, and attribute-based classification \cite{Alpher50}, etc.

As mentioned above, pedestrian attribute recognition has been widely used for person Re-ID.
On the one hand, attribute recognition focuses on local aspects of a person while person re-ID usually extracts global representations.
The joint training of attribute-identity and ID-identity is beneficial to complement each other and improve the performance of the both tasks.
On the other hand, object pedestrians may have the same attributes. For example, the thieves who are looked for in the surveillance system have common attributes, such as having a backpack, wearing a hat,
wearing sneakers and so on. Therefore, adding attribute recognition to the pedestrian recognition is also conducive to searching target pedestrians faster and better in the actual surveillance system.
All in all, considering their similarity and difference, this paper proposes a neural network that learns a re-ID embedding and predicts the pedestrian attributes simultaneously.

Although the combination of the two tasks can complement each other to a certain extent, there are still some problems, containing low pixels, variations in viewpoints and illumination,  background clutter, attribute recognition  at far distance, all making them difficult to tackle. Most methods \cite{Alpher41, Alpher51, Alpher01, Alpher08} focus on extracting features which are discriminative and robust to these variations and train a separate classifier for each attribute.
In fact, attributes of pedestrian always show semantic correlation by which they can be associated. Extracting correlation information among features is also beneficial to address these difficulties.

To obtain the context of the attributes, literature \cite{Alpher33} proposes a sequence to sequence framework, which maps both the manual division images and attributes into sequences with the CNN-RNN based encoder-decoder structure.
However, the results may influenced by different manual division and attribute orders due to the weak alignment ability of RNN, which is somewhat unreasonable.
While this paper not only maps the attribute labels into number sequences through the mapping table,
but also encodes the images into sequences by the neural networks without division. Moreover, connectionist temporal classification (CTC) loss \cite{Alpher40} is utilized to strengthen the encoding ability of networks,
which can align sequences automatically so that it can avoid being affected by attribute order in the mapping table.
According to our knowledge,
this is the first time that CTC loss is used in attribute recognition task and has achieved good performance.

\begin{figure}[t]
\begin{center}
\includegraphics[width=0.9\linewidth]{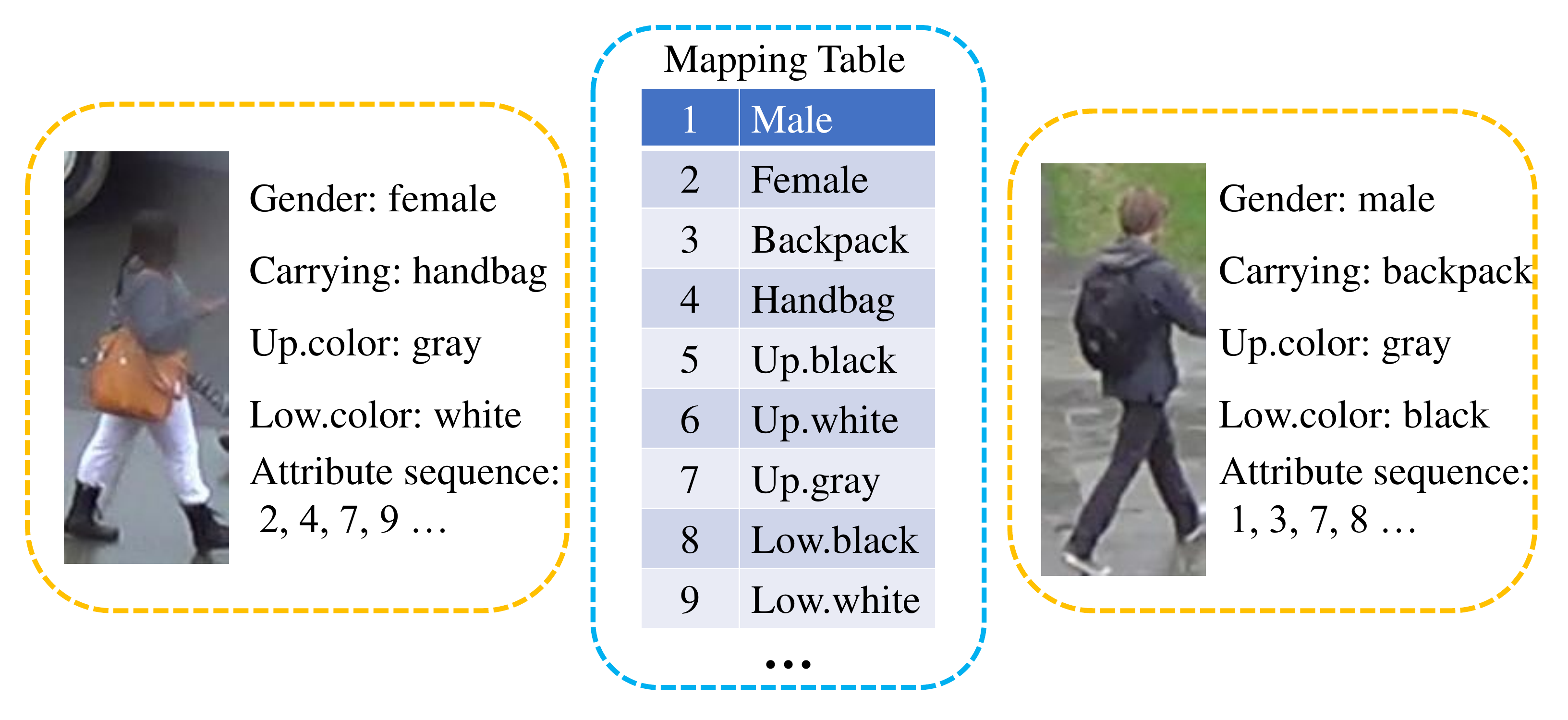}
\end{center}
   \caption{In this paper, a person's appearance can be described by appearance factors of mapped attribute sequences with the mapping table. `Up.color' and `Low.color' denote
the color of upper-body clothing and the color of lower-body clothing, resp. }
\label{fig:1}
\end{figure}

Recently, some deep attention methods have been proposed to pay attention to more significant local regions in attribute recognition tasks.
However, when the saliency region is not located accurately, it will bring quadratic error. The attention model used in this paper is the Transformer \cite{Alpher20},
which has been successfully used in translation tasks. We utilize its language learning ability to decode sequences,
which can not only align image and attribute sequences but also better learn semantic features from attributes. Thus, the orders of attribute in the sequences will not affect the performances.
And it can predict the entire sequence at a time, because it allows our neural network to predict labels at any time interval.

To sum up, this paper simultaneously learns a discriminative neural network embedding for an attribute prediction and re-ID model.
And it combines attention model and CTC loss forming a joint CTC-Attention model (JCM) to implement attribute recognition task,
which can predict multiple attribute values with arbitrary length at a time avoiding the influence of attribute order in the mapping table. The main contributions are summarized as follows:
(1) This paper maps all attributes into a sequence of numbers and encodes the images into sequences for unified learning, and to our best knowledge, it is the first time to use the CTC loss function to learn the context among attributes.
(2) JCM employs the attention model to realize the alignment operation of the input and output sequences and the extraction of the semantic information from the sequences.
(3) In-depth experiments are operated to analyze various aspects of our method. Moreover, JCM yields competitive accuracy in attribute recognition and demonstrating some improvement in Re-ID.


\section{Related Works}
This part briefly reviews the attribute recognition task and Re-ID task respectively.
\subsection{Attributes recognition}

Attribute recognition methods include hand-crafted feature methods and deep feature methods.
Besides, this task has also been extensively studied with methods falling into two aspects: independent multiple attributes and dependent multiple attributes.

Hand-crafted feature methods usually extract color and texture information from the images. In this kind of methods,
for independent multiple attributes, most methods  \cite{Alpher41,Alpher42,Alpher43, Alpher51, Alpher52} train a separate classifier for each attribute.
In \cite{Alpher41, Alpher51, Alpher52}, SVM and low-level descriptors are used to train attribute detectors, and the attributes are integrated in several metric learning methods.
While for dependent multiple attributes, they consider inter-attribute correlation as an extra information for improving recognition performance.
Graph model based methods \cite{Alpher44, Alpher27, Alpher45} capture attribute co-occurrence likelihoods by using conditional random field or Markov random field.
But existing graph models are high computational cost when dealing with a large set of attributes.

Recently, deep feature methods \cite{Alpher01,Alpher02, Alpher07,Alpher08,Alpher16, Alpher33, Alpher35, Alpher46, Alpher47,Alpher48,Alpher49} have been adopted in pedestrian attribute recognition task to learn more expressive representations,
which significantly improve the performance of pedestrian attribute recognition.
In deep methods,  for independent multiple attributes,  literatures \cite{Alpher01,Alpher08,Alpher16,Alpher46} integrate separate attribute classification losses for each attribute.
Khamis et al. \cite{Alpher16} propose to jointly optimize the triplet loss for re-ID and the attribute classification loss. DeepMar model \cite{Alpher46} utilizes the prior knowledge in the object topology for attribute recognition,
and designs a weighted sigmoid cross entropy loss to deal with the data imbalance problem.
For dependent multiple attributes, approaches \cite{Alpher02,Alpher33} aim to discover the interdependency and correlation among attributes with long short-term memory (LSTM) model.
A CNN-RNN based encoder-decoder framework is proposed in \cite{Alpher33}. Literature \cite{Alpher02} predicts attributes group by group via mining both semantic and spatial correlations in attribute groups.

However, literature \cite{Alpher33} predicts attributes one by one, which is very expensive in computation.
It requires coding the divided images and manually arranging the order of images and attributes.
While this paper uses JCM to automatically align the sequence of attributes without dividing the image, and it can output arbitrary length sequence.

Attention models for attribute recognition most focus on spatial attention. \cite{Alpher03,Alpher04,Alpher05,Alpher06,Alpher35} locate the corresponding attribute space location and extract the feature to identify the attribute of the part.
Literature \cite{Alpher35} is proposed to avoid the negative effect of irrelevant image region. The attention model in our method does not consider the spatial relations,
but focuses on the semantic information among attributes, so as to achieve the alignment of attribute sequences.

\subsection{Person re-identification}

Person re-identification is to determine whether the pedestrians from different cameras are the same pedestrians.
However, there are many factors making it difficult to tackle, including the low resolution of the videos, background clutter and the pedestrian occlusion, etc.
Moreover, the same pedestrians may have different viewpoints and illumination in different videos which are taken from the non-overlapping cameras, and different pedestrians may have similar appearances and gaits.
Thus, the intra-class variations may be larger than the inter-class variations. To solve these problems,
the key steps of person Re-ID are extracting discriminative features and accurately evaluating the similarities between features.
The method of person Re-ID is also divided into traditional and deep methods.

In traditional methods, for feature extraction, most approaches extract the hand-crafted features \cite{Alpher53,Alpher54,Alpher55},
which should be discriminative and robust to the variance among different cameras. The hand-crafted features contain the appearance features and the spatial-temporal features,
such as color, texture, edge, optical flow, etc. Literature \cite{Alpher55} exploits the periodicity exhibited by a walking person to generate a spatiotemporal body-action model,
which consists of a series of body-action units corresponding to certain action primitives of certain body parts. For similarities evaluating, approaches usually map the features into a metric space \cite{Alpher54,Alpher56,Alpher57},
where the intra-class variations are smaller and the inter-class variations are larger. Then they calculate the distances between the probe and the gallery in the metric space.
In \cite{Alpher54}, You et al. propose a top-push distance metric learning which optimizes the top-rank matching.

In deep methods, for feature extraction,
most approaches extract the deep features which include the spatial features extracted from convolutional neural network (CNN) and the correlation features extracted from recurrent neural network (RNN) \cite{Alpher07,Alpher09,Alpher34,Alpher58,Alpher59}.
Literature \cite{Alpher07} proposes a multi-level factorization net (MLFN) that factorizes the visual appearance of a person into latent discriminative factors at multiple semantic levels.
\cite{Alpher09} integrates human semantic parsing to extract local features from human body parts. Similarities evaluating is realized by optimizing the networks with different loss functions.
Commonly used loss functions contain classification loss and verification loss. Classification loss is calculated with the probability of the current image belonging to each class \cite{Alpher01}.
Verification loss includes Siamese loss and Triplet loss and so on \cite{Alpher17,Alpher18,Alpher19}. Liu et al. \cite{Alpher17} propose an end-to-end Siamese network that extracts both appearance and motion context information.
At present,
it is commonly used to optimize the classification loss function and verification loss function to train the network simultaneously.

\begin{figure*}[t]
\begin{center}
\includegraphics[width=0.7\linewidth]{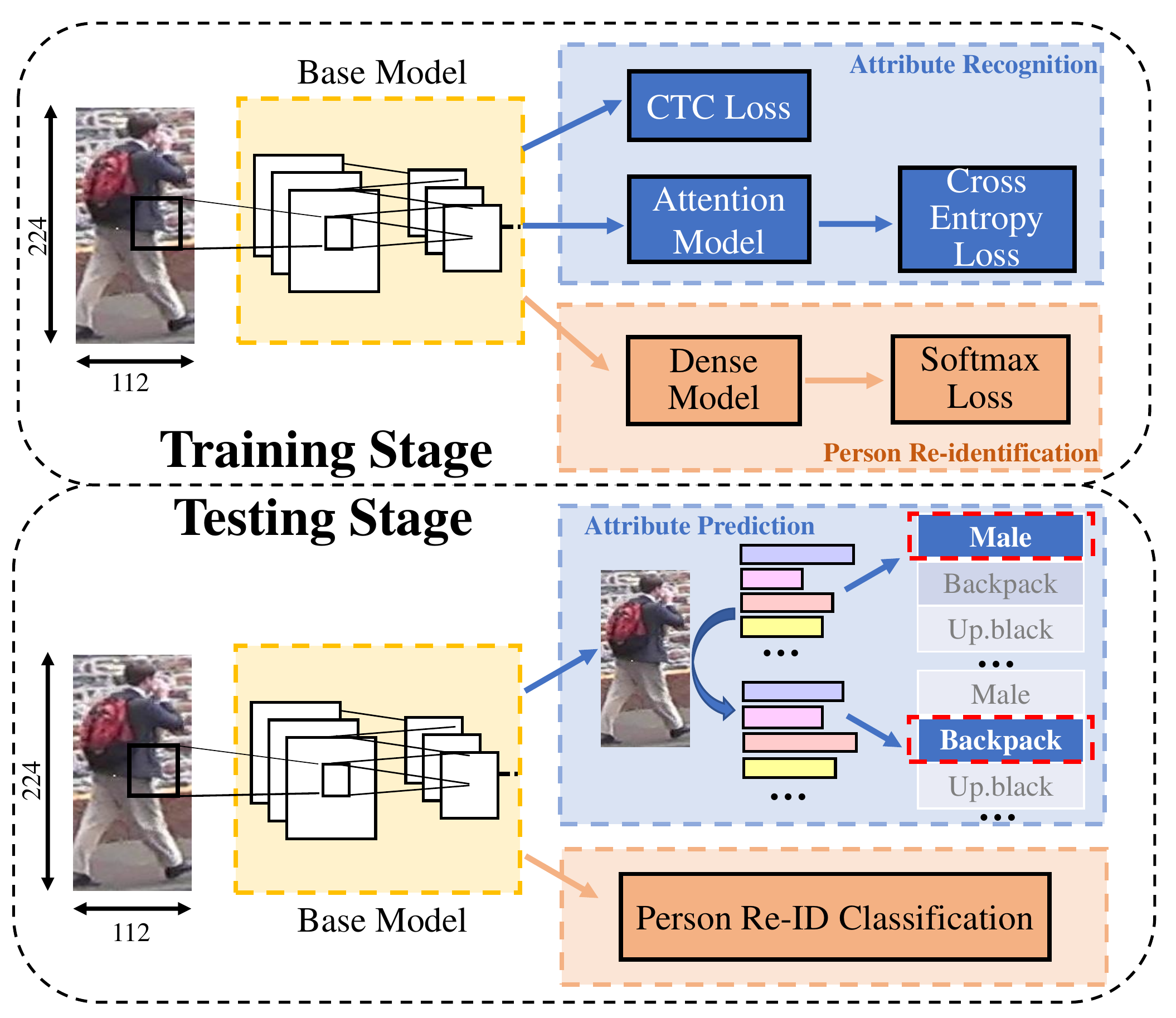}
\end{center}
   \caption{An overview of the proposed network.}
\label{fig:2}
\end{figure*}

\section{Methods}
\subsection{Architecture}

The procedure of the JCM consists of two parts: training stage and
testing stage depicted in the Fig. \ref{fig:2}. The JCM consists of a base model, a dense model and an attention model.
The input image is fixed to $224\times112$ size.

In detail, when training the proposed network, for attribute recognition stream, it encodes the images with the base model,
which includes convolution neural network (CNN) and recurrent neural network (RNN). Then the output of the base model is feed into CTC loss and attention model respectively to train attribute recognition task.
For person Re-ID stream, it extracts features from images with the base model. Then the dimension of the feature is reduced to the number of ID by dense model.
And it adopts softmax loss to train the person Re-ID task. The specific training procedure will be described later.

When testing, the proposed network simultaneously predicts the identity and a set of attributes.  To predict the attribute sequence for attribute recognition task, it decodes the encoded features by the attention model with beam search.
At the same time, it extracts the features from the CNN  in base model to classify pedestrians for person re-ID task.

\subsection{Base model \& Dense model}
Base model is composed of CNN and bidirectional RNN. In recent years, Resnet50 \cite{Alpher13} has been widely used in classification tasks for extracting features.
Therefore, our CNN adopts the same structure as Resnet50 with different parameters. Our RNN contains two layers. The specific parameters of base model are shown in TABLE \ref{tab1}:

It can be seen from TABLE \ref{tab1} that a convolution block consists of three convolution layers.
Four stages, including conv2\_x, conv3\_x, conv4\_x and conv5\_x, all consist of several convolution blocks.
The stride of the first convolution layer of the first convolution block in each stage is set to [2,1], and that of the remaining convolution layers is set to 1.

Dense model first reshapes the output of CNN in base model to single row vector.
Then two fully connected layers $FC_{0}$, $FC_{1}$ are applied, whose outputs are $c = [c_{1}, c_{2},\cdots, c_{1024}]$,
 $z = [z_{1}, z_{2},\cdots, z_{N}] \in R^{N}$ respectively,  $N$ is the number of ID. So, the predicted probability of each ID label $n$, $n \in 1,2,\cdots,N$, is calculated as:
\begin{equation}
p(n|z)=\frac{\exp(z_{n})}{\sum_{i=1}^{N}\exp(z_{i})}
\label{Eq.1}
\end{equation}
The cross entropy loss of ID classification can be formulated as below:
\begin{equation}
L_{ID} = -\sum_{n=1}^{N} \log(p(n))q(n)
\label{Eq.2}
\end{equation}

Let $g$ be the ground-truth ID label, so that q(g) = 1 and q(n) = 0 for all $n\neq g$. In this case,
minimizing the cross entropy loss is equivalent to maximizing the possibility of being assigned to the ground-truth class.

\begin{table}
\begin{center}

\begin{tabular}{|c|c|c|}
\hline
Layer name & Output size & Patameters \\
\hline\hline
conv\_1 & $112\times 56$ & $7\times 7$, 64, stride 2 \\
\hline
maxpool\_1 & $56\times 28$ & $ 3\times 3$, 64, stride 2     \\
\hline
conv2\_x & $28\times 28$ &$\begin{Bmatrix} 1\times1,64\\ 3\times3,64\\ 1\times1,256\\\end{Bmatrix}\quad\times 3$ \\
\hline
conv3\_x & $14\times 28$ & $\begin{Bmatrix} 1\times1,128\\ 3\times3,128\\ 1\times1,512\\\end{Bmatrix}\quad\times 4$ \\
\hline
conv4\_x & $7\times 28$  & $\begin{Bmatrix} 1\times1,256\\ 3\times3,256\\ 1\times1,1024\\\end{Bmatrix}\quad\times 6$\\
\hline
conv5\_x & $4\times 28$  & $\begin{Bmatrix} 1\times1,512\\ 3\times3,512\\ 1\times1,2048\\\end{Bmatrix}\quad\times 3$ \\
\hline
maxpool\_2 & $1\times 28$ & $ 3\times 1$, 2048, stride [3,1]     \\
\hline
rnn\_1 & $1\times 28$ &   rnn size 1024    \\
\hline
rnn\_2 & $1\times 28$ &   rnn size 512    \\
\hline
\end{tabular}
\end{center}
\caption{The specific parameters of base model.}
\label{tab1}
\end{table}

\subsection{CTC}

In order to obtain relationship among attributes,
we map semantic attribute labels to corresponding number labels through mapping table.
As shown in Fig. \ref{fig:1}, multiple attributes of each person are represented by the attribute sequence $y$, $y = [y_{1}, \cdots, y_{t}, \cdots, y_{U}], y_{t} \in 1,2,\cdots,K$,
$K$ is the number of attributes, $U$ is the length of the sequence, each person's sequence length can be different.

Given $x$ is the features which are encoded by the base model with length $T$ and dimension $D$, $x = [x_{1}, x_{2},\cdots, x_{T}]$,  $T$ equals to 28 and $D$ equals to 1024 in this paper. We feed it into CTC loss to improve the coding capability of network.
The key idea of CTC is to use intermediate label representation $\pi = [\pi_{1}, \pi_{2},\cdots, \pi_{T}]$, $\pi$ belongs to the probability distribution over all possible label sequences $\Phi(y')$,
where $y'$ is a modified label sequence of $y$,
which is made by inserting the blank symbols between each label and the beginning and the ending of the sequences (i.e., y = (1;5;8); $y¡¯$¡¯=(-;1;-;5;-;8;-)).
CTC trains the model to maximize $P(y|x)$:
\begin{equation}
P(y|x) = -\sum_{\pi\in\Phi(y')} P(\pi|x)
\label{Eq.3}
\end{equation}

CTC is applied on the top of RNN.
Each RNN output unit is interpreted as the probability of observing the corresponding label at particular time.
The probability of label sequence $P(\pi|x)$ is modeled as being conditionally independent by the product of the network outputs:
\begin{equation}
P(\pi|x) \approx \prod_{t=1}^{T} P(\pi_{t}|x) = \prod_{t=1}^{T}q_{t}(\pi_{t})
\label{Eq.4}
\end{equation}
where $q_{t}(\pi_{t})$ denotes the softmax activation of $\pi_{t}$ label in RNN output layer $q$ at time $t$.

The CTC loss to be minimized is defined as the negative log likelihood of the ground truth character sequence $y$, i.e.
\begin{equation}
L_{CTC} \triangleq -\ln P(y|x)
\label{Eq.5}
\end{equation}

\subsection{Attention model}

Attention mechanisms have become an integral part in various tasks. This paper pays attention to applying the mechanism on the compelling sequence modeling,
allowing modeling of dependencies without regard to their distance in the input or output sequences, so that the encoded image sequences outputted by base model and the attribute sequences can be aligned and the semantic relationship of each attribute in the attribute sequence can be better learned.
To be specific, this paper adopts the attention model, the Transformer, described in \cite{Alpher20} to achieve it.

The Transformer architecture adopts encoder-decoder structure. Both encoder and decoder are composed of a stack of $N = 6$ identical layers. Each layer contains several sub-layers,
and it employs residual connections around each of the sub-layers. To facilitate these residual connections, all sub-layers in the model, as well as the embedding layers, produce outputs of dimension $d_{model}$ = 1024.
Among these sub-layers, attention sub-layers consist of $h$ attention layers running in parallel. Thus, it is especially suitable for the alignment of long attribute sequence. In this paper, we employ $h = 8$.

In this paper, the inputs of the encoder are the encoded features outputted by the base model, with the max length of 28.
Since the images have been encoded into 1024 dimensions, the encoder in the attention model no longer needs the embedding layers. The inputs of the decoder are the attribute sequence labels.
The length of the decoder sequence is increased to 28 by adding `0' in the tail of the sequence. Then all the elements in the sequence are moved one bit to the right and  `100' is added to the first position of the sequence as the starting marker.

After the attention model, the fully connection layer F is connected.
The output of the F layer is $f = [f_{1}, f_{2},\cdots, f_{K}] \in R^{K}$.
So the predicted probability of each attribute label $k$, $k \in 1,2,\cdots,K$, is denoted as $p(k|f)$:
\begin{equation}
p(k|f)=\frac{\exp(f_{k})}{\sum_{i=1}^{K}\exp(f_{i})}
\label{Eq.8}
\end{equation}
The cross entropy loss of attribute classification can be formulated as below, which is the same as Eq. \ref{Eq.2}:
\begin{equation}
L_{AT} = -\sum_{k=1}^{K} \log(p(k))q(k)
\label{Eq.9}
\end{equation}

By using a CTC loss function, a multi-attribute classification loss function and an identity classification loss function, the network is trained to predict attribute and identity labels.
Here the final loss function is defined as:
\begin{equation}
L =  \lambda L_{ID} +  L_{CTC} + L_{AT}
\label{Eq.7}
\end{equation}

Parameter $\lambda$ balances the contribution of the three losses and is determined on a validation set of Market-1501.

\section{Experiments}
\subsection{Datasets and Evaluation Protocol}

$\textbf{Market-1501 attribute dataset}$ \cite{Alpher01} is an extension of Market-1501 dataset \cite{Alpher10} with person attribute annotations.
The dataset contains 32688 images of 1501 identities. Among them 751 and 750 identities are used as the training and testing set respectively.
It contains 12 different types of annotated attributes, including 10 binary attribute (such as gender, hair length, and sleeve length) and 2 multi-class attributes, i.e. colors of upper and lower body clothing.

$\textbf{Duke attribute dataset}$ \cite{Alpher01} is an extension of DukeMTMC-reID dataset \cite{Alpher11} with person attribute annotations, which contains 1812 identities captured under 8 cameras.
Training and testing set both have 702 identities with 16522 training images and 17661 testing gallery images respectively. It is annotated with 8 binary pedestrian attributes such as wearing a hat,
and wearing boots, and 2 multi-class attributes.

$\textbf{The PETA dataset}$ \cite{Alpher62} is a collection of 10 person surveillance datasets and consists of 19000 cropped images along with 61 binary and 5 multi-value attributes.
This paper only employs 8 datasets which have more than two images per ID. They include 13549 images with 3268 identities. We randomly divide the dataset into training set and testing set by half, namely,
there are 1634 identities in the training set and the remaining 1634 identities are constructed as the testing set.
The experiments are repeated 10 times to get average results.

$\textbf{Evaluation metrics:}$ For the attribute recognition task, we evaluate the classification accuracy for each attribute (24, 21 and 66 attributes for Market-1501, DukeMTMC-reID and PETA datasets, respectively).
Besides,  the mean accuracy of all attributes is calculated. For Market-1501, DukeMTMC-reID, the gallery images are used as the testing set. For PETA, the testing images are used as the testing set.

For the person re-ID task, the cumulative matching characteristic (CMC) curve and the mean average precision (mAP) are adopted.
For each query, its average precision (AP) is computed from its precision-recall curve. Then mAP is the mean value of average precisions across all queries.
We report the cumulated matching result at selected Rank-1 instead of plotting the actual curves.

\subsection{Implementation Details}

When base model is  pretrained on the Imagenet2012 dataset \cite{Alpher60},
the number of epochs is set to 300. When the whole model is trained on the pedestrian attribute datasets, the number of epochs is set to 200.
For both pretraining and training, the batch size is set to 64. The learning rate is 1e-4 and the decay-rate is 0.9.
The adaptive moment estimation (ADAM) \cite{Alpher21} is implemented in each mini-batch to update the parameters.

Data Augmentation: Left-right flip augmentation is used during training. No data augmentation is used for testing.

\subsection{Evaluation of Attribute Recognition}

\begin{figure}[t]
\begin{center}
\includegraphics[width=0.9\linewidth]{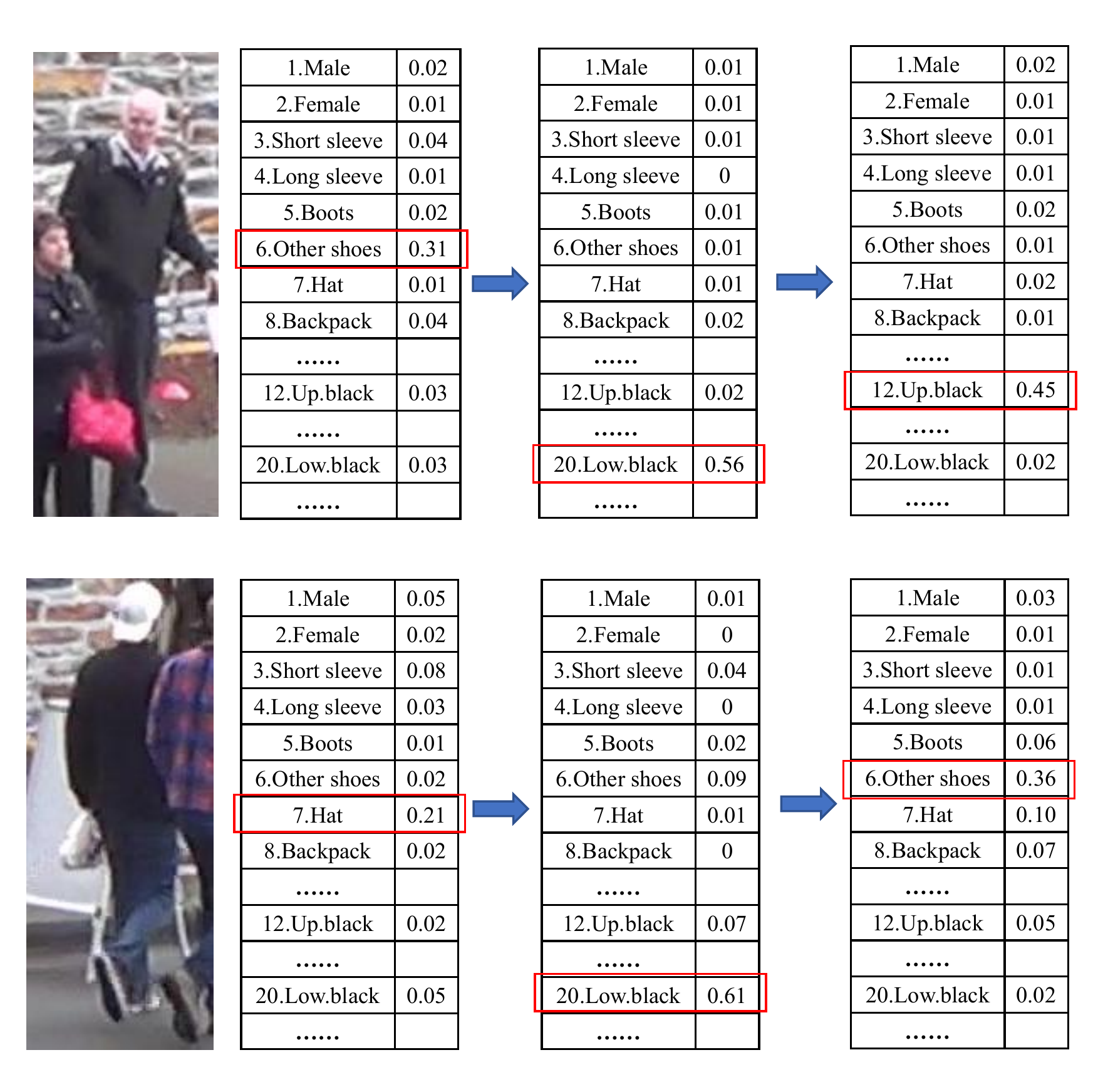}
\end{center}
   \caption{Examples for person attribute recognition. Red bounding boxes indicate the selected attributes with the max probabilities.}
\label{fig:5}
\end{figure}

$\textbf{Results on Market-1501 Dataset:}$ This paper first evaluates our method against the state-of-the-arts on Market-1501 dataset for attribute recognition. As seen from TABLE \ref{tab2},
our method can achieve the mean accuracy of 89.7\% over 12 pedestrian attributes, outperforming the current best method AWMDN 1.2\%.
From each attribute classification result, 5 attribute recognition rates of our method are the best.
Not only that, it can seen that the 12 attributes recognition rates in our method differ little, the best recognition rate in `hat' is 14.7\% higher than the worst in `hair',
while the best recognition rate of method \cite{Alpher23} in `hat' is 29.8\% higher than the worst in `bag', which shows that our attribute prediction method is more uniform and stable.

\begin{table*}
\begin{center}

\setlength{\tabcolsep}{2.7mm}{
\begin{tabular}{|c|c|c|c|c|c|c|c|c|c|c|c|c|c|}
\hline
Method & gender &age &hair  &L.slv	&L.low	&S.clth	&B.pack	&H.bag	&bag	&hat	&C.up	&C.low	&mA \\
\hline\hline
APR \cite{Alpher01}	&85.8	&87.7	&83.5	&94.1	&92.6	&91.4	&86.3	&76.0	&91.3	&88.2	&89.7	&91.3	&88.2 \\
Sun et al.\cite{Alpher23}	&88.9	&84.8	&78.3	&93.5	&92.1	&84.8	&85.5	&88.4	&67.3	&97.1	&87.5	&87.2	&86.3 \\
AWMDN\cite{Alpher24}	&-	&-	&-	&-	&-	&-	&-	&-	&-	&-	&-	&-	&\textcolor{blue}{88.5}\\
MLFN \cite{Alpher07} &-	&-	&-	&-	&-	&-	&-	&-	&-	&-	&-	&-	&85.3\\
PANDA\cite{Alpher37} &-	&-	&-	&-	&-	&-	&-	&-	&-	&-	&-	&-	&86.8\\
JCM (our)	&$\textbf{89.7}$	&$\textbf{87.4}$	&$\textbf{82.5}$	&$\textbf{93.7}$	&$\textbf{93.3}$	&$\textbf{89.2}$	&$\textbf{85.2}$	&$\textbf{86.2}$	&$\textbf{86.9}$	
&$\textbf{97.2}$	&$\textbf{92.4}$	&$\textbf{93.1}$	&\textcolor{red}{$\textbf{89.7}$}\\

\hline
\end{tabular}}
\end{center}
\caption{Attribute recognition accuracy on Market-1501. `L.slv', `L.low', `S.clth',
`B.pack', `H.bag', `C.up', `C.low' denote length of sleeve, length of lower-body clothing, style of clothing, backpack, handbag, color of
upper-body clothing and color of lower-body clothing, resp. }
\label{tab2}
\end{table*}

$\textbf{Results on DukeMTMC-reID Dataset:}$ TABLE \ref{tab3} shows the comparisons between our method and state of the arts on the DukeMTMC-reID dataset.
It can be observed that our method is better than all comparison methods. More concretely, we can achieve the mean accuracy of 89.0\% over 10 pedestrian attributes,
which is 0.7\% higher than the best result 88.3 \% reported  so far  in \cite{Alpher23}. Similarly, the recognition rates of 10 attributes in our method are close.

\begin{table*}
\begin{center}
\setlength{\tabcolsep}{3.85mm}{
\begin{tabular}{|c|c|c|c|c|c|c|c|c|c|c|c|}
\hline
Method & gender &hat	&boots	&L.slv	&B.pack	&H.bag	&bag	&C.shoes	&C.up	&C.low	&mA \\
\hline\hline
APR\cite{Alpher01}	&81.6	&86.3	&86.0	&88.6	&74.7	&93.6	&82.1	&90.0	&99.5	&81.9	&86.4 \\
SVM \cite{Alpher22}	&77.0	&82.2	&82.5	&87.6	&69.6	&93.6	&83.0	&90.0	&70.9	&68.5	&80.5 \\
Sun et al.\cite{Alpher23}	&88.9	&83.0	&80.1	&93.6	&87.0	&89.6	&91.6	&83.7	&93.9	&91.8	&\textcolor{blue}{88.3}\\
AWMDN\cite{Alpher24}	&-	&-	&-	&-	&-	&-	&-	&-	&-	&-		&87.5\\
MLFN \cite{Alpher07} &-	&-	&-	&-	&-	&-	&-	&-	&-	&-		&87.5\\
PANDA\cite{Alpher37} &-	&-	&-	&-	&-	&-	&-	&-	&-	&-		&85.9\\
JCM (our)	&$\textbf{87.4}$	&$\textbf{83.3}$	&$\textbf{89.6}$	&$\textbf{88.3}$	&$\textbf{89.0}$	
&$\textbf{92.4}$	&$\textbf{87.9}$ &$\textbf{87.1}$	&$\textbf{92.9}$	&$\textbf{92.1}$	&\textcolor{red}{$\textbf{89.0}$}\\

\hline
\end{tabular}}
\end{center}
\caption{Attribute recognition accuracy on DukeMTMC-reID. `C.shoes' denotes color of shoes, and the other notations are the same with
TABLE \ref{tab2}. }
\label{tab3}
\end{table*}

$\textbf{Results on PETA Dataset:}$
We divide the 65 attributes into 12 categories. The average recognition rate of each class of attributes is calculated, and the mean accuracy of 12 classes of attributes is computed.
Because there are many attributes in PETA dataset, most methods only select partial attributes for attribute recognition task.
Or some methods use all attributes for attribute recognition, but only list the accuracy of partial attributes.
Nevertheless, the performance of the method can still be judged by mean accuracy. The TABLE \ref{tab4} shows that our method can achieve the mean accuracy of 91.3\% over 12 pedestrian attributes,
and its performance is superior to all other methods.

\begin{table*}
\begin{center}
\resizebox{\textwidth}{!}{
\begin{tabular}{|c|c|c|c|c|c|c|c|c|c|c|c|c|c|}
\hline
Method & gender &age &C.hair	&L.hair	&C.shoes	&carrying	&accessory	&T.up	&T.low	&T.shoes	&C.up	&C.low	&mA \\
\hline\hline
Chen et al.\cite{Alpher25}	&88.9	&91.3	&-	&91.3	&-	&87.4	&95.1	&91.5	&89.1	&85.6	&-	&-	&\textcolor{blue}{90.0} \\
MLCNN \cite{Alpher26}	&84.3	&87.8	&-	&88.1	&-	&83.7	&93.1	&88.8	&85.2	&80.9	&-	&-	&86.5 \\
Super-fine\cite{Alpher30}	&93.1	&86.8	&-	&89.8	&-	&81.9	&83.5	&77.3	&83.7	&77.3	&-	&-	&84.2\\
MAResNet\cite{Alpher61}	&76.6	&80.3	&-	&75.9	&-	&73.0	&79.6	&73.9	&75.2	&73.3	&-	&-	&76.0\\
DeepMar\cite{Alpher46} &89.9     &87.2  &-  &88.9   &-   &82.4  &85.9   &83.8   &79.1   &78.3   &-	&-	&84.4\\
ACRN \cite{Alpher28}	&-	&-	&-	&-	&-	&-	&-	&-	&-	&-	&-	&-	&84.6\\
WPAL\cite{Alpher29}	&-	&-	&-	&-	&-	&-	&-	&-	&-	&-	&-	&-	&85.5\\
Sarafianos et al.\cite{Alpher31}	&-	&-	&-	&-	&-	&-	&-	&-	&-	&-	&-	&-	&84.6\\
GRL\cite{Alpher32}	&-	&-	&-	&-	&-	&-	&-	&-	&-	&-	&-	&-	&85.7\\
JRL\cite{Alpher33}	&-	&-	&-	&-	&-	&-	&-	&-	&-	&-	&-	&-	&85.7\\
JCM (our)	&$\textbf{89.3}$	&$\textbf{84.5}$	&$\textbf{92.9}$	&$\textbf{92.7}$	&$\textbf{92.3}$	&$\textbf{91.8}$	
&$\textbf{90.2}$	&$\textbf{89.7}$	&$\textbf{90.4}$	&$\textbf{88.2}$	&$\textbf{92.9}$ &$\textbf{93.5}$	&\textcolor{red}{$\textbf{90.7}$}\\

\hline
\end{tabular}}
\end{center}
\caption{Attribute recognition accuracy on PETA. `C.hair', `L.hair', `T.up', `T.low' and `T.shoes' denote color of hair, length of hair, type of upper-body clothing, type of lower-body clothing and type of shoes, resp. And the other notations are the same with TABLE \ref{tab2}.}
\label{tab4}
\end{table*}

$\textbf{Examples of attribute recognition results:}$ We show two examples of attribute prediction in Fig. \ref{fig:5}.
The weights of first three steps which are produced by the attention model are listed, which proves that this model can achieve aligning the image and the attribute sequences and processing the semantic information well.

\subsection{Evaluation of Person Re-ID}

The experiments are conducted on two person Re-ID datasets. Although this paper focuses on attribute recognition, it can still achieve good performance in person Re-ID. As shown in TABLE \ref{tab5},
our method achieves 93.1\% Rank-1 matching rate on the Market-1501 dataset with utilizing re-ranking \cite{Alpher12}, which is better than all works which combines attributes and person recognition, including APR, MLFN, ACRN, HP-net. But it is slightly lower than HA-CNN which specializes in person Re-ID.
Moreover, it can be observed that our method can beat the comparative method by a large margin on the PETA dataset.
Since PETA is a collection of several datasets, including both single shot and multi-shot, re-ranking is not applied on this dataset.

\begin{table}
\begin{center}
\setlength{\tabcolsep}{1.5mm}{
\begin{tabular}{|c|c|c|c|}
\hline
Dataset &Method & Rank-1 &mAP  \\
\hline\hline

\multirow{8}*{Market-1501} &APR \cite{Alpher01}	&84.3	&64.7 \\
&MLFN \cite{Alpher07}	&92.3	&82.4 \\
&HA-CNN\cite{Alpher34}	&\textcolor{red}{93.8}	&\textcolor{red}{82.8}\\
&ACRN \cite{Alpher28}	&83.6	&62.6\\
&HP-net\cite{Alpher35}	&76.9	&-\\
&JCM (D, 1024)	&$\textbf{84.9}$	&$\textbf{75.7}$\\
&JCM (D, 57344)	&$\textbf{91.3}$	&$\textbf{81.2}$\\
&JCM (D, 57344) + Re-ranking & \textcolor{blue}{$\textbf{93.1}$}	&\textcolor{blue}{$\textbf{82.5}$}\\
\hline
\multirow{3}*{PETA} &DC-ML\cite{Alpher36}	&50.1 &-\\
&JCM (D, 1024)	&$\textbf{57.3}$	&-\\
&JCM (D, 57344) & \textcolor{red}{$\textbf{62.5}$}	&-\\
\hline
\end{tabular}}
\end{center}
\caption{Comparison with the person Re-ID state of the art on Market-1501 and PETA datasets.
Rank-1 accuracy (\%) and mAP (\%) are shown. `D' represents the dimension of the features. }
\label{tab5}
\end{table}

\subsection{Ablation Studies}
\subsubsection{Parameter validation}
\begin{figure}[t]
\begin{center}
\includegraphics[width=0.9\linewidth]{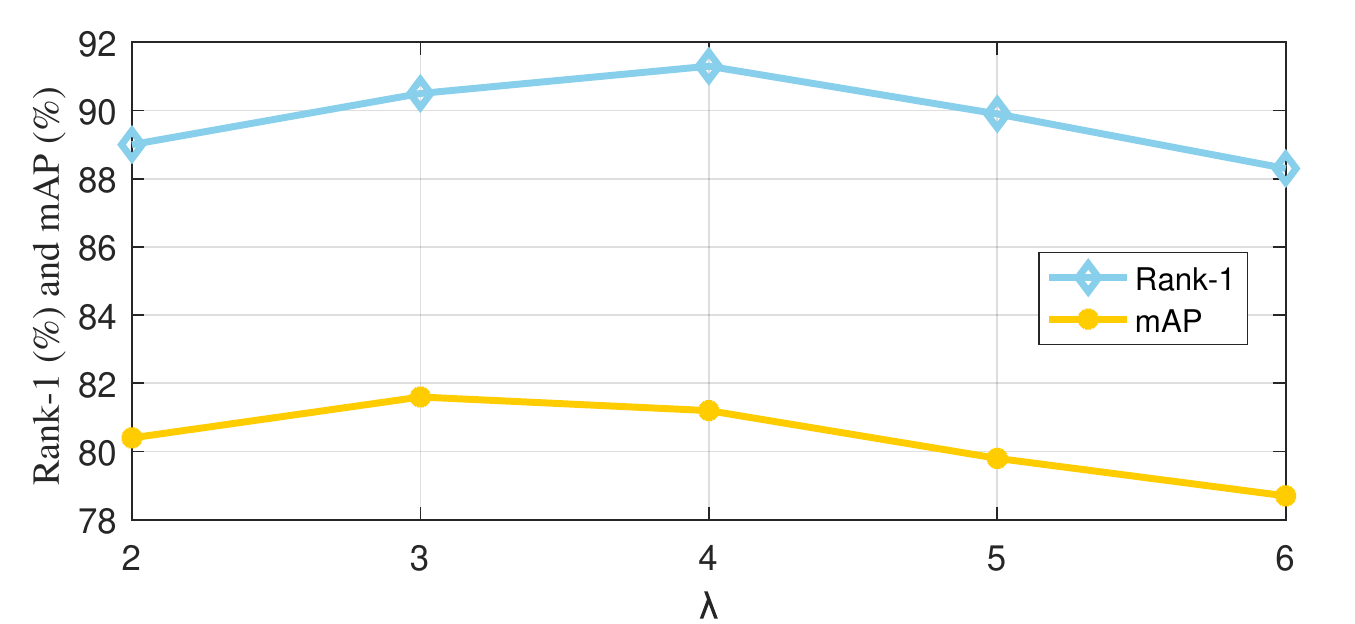}
\end{center}
   \caption{Person Re-ID validation results of parameter $\lambda$ on Market-1501 dataset.}
\label{fig:6}
\end{figure}
This paper  shows the re-ID validation results of parameter $\lambda$ which is a key parameter balancing the contribution of re-ID and attribute recognition (Eq. \ref{Eq.7}).
Re-ID results on the validation set of Market-1501 are presented in Fig. \ref{fig:6}. From the mAP and Rank-1 results, it can be observed that both curves increase first then decrease. When $\lambda$ = 4 ,
a relatively finer re-ID performance can be achieved. Therefore,  $\lambda$  is set to 4 on three datasets and in the later experiments.

\subsubsection{The effect of joint learning}
We compare the use of joint training with performing attribute recognition and person re-identification tasks respectively on the Market-1501 dataset. The results are shown in the TABLE \ref{tab6}.
It can be seen that the recognition rates of attribute recognition in our method are higher than those of the method only performing attribute recognition, so does the Re-ID.
Based on the experiments, it can be concluded that attribute recognition can improve re-ID, and vice versa.

\begin{table}
\begin{center}
\setlength{\tabcolsep}{2mm}{
\begin{tabular}{|c|c|c|c|}
\hline
Dataset &Method & Rank-1 &mA  \\
\hline\hline
\multirow{3}*{Market-1501}&JCM w/o person Re-ID	&-	&88.8\\
&JCM w/o attribute recognition 	&89.0	&- \\
&JCM (our)	&$\textbf{91.3}$	&$\textbf{89.7}$\\
\hline
\multirow{2}*{TownCenter}&Independent Training	&83.9 &90.1	\\
&Hybrid  Training  	&$\textbf{87.7}$ &\textbf{91.8}	\\
\hline
\end{tabular}}
\end{center}
\caption{The results of performing attribute recognition and person Re-ID tasks
respectively on the Market-1501 dataset, and the results of different training strategies on TownCenter dataset.
Rank-1 accuracy (\%) and mA (\%) are shown. }
\label{tab6}
\end{table}

\subsubsection{The features extracted from different layers}
In order to evaluate the performance of features extracted from different layers in the network on person Re-ID, in addition to extracting features from CNN in base model which are $28 \times 2048$ = 57344 dimension, we extract the representations from the $FC_{0}$ layer in dense model, which are 1024 dimensions.
The experimental results  on the Market-1501 and PETA datasets are listed in the TABLE \ref{tab5}.
As we can see from the results, the features extracted from CNN in base model have much better performance than those of dense model on both datasets.




\subsubsection{The effect of hybrid training}
As described above, 8 datasets in PETA dataset are employed in this paper. And our method randomly selects half samples as training sets from the combined 8 datasets.
To verify the effectiveness of this hybrid training, we choose one dataset from PETA dataset for independent training.
To be fair, the largest dataset TownCenter dataset is chosen, which contains 6967 images with 231 IDs taken in the outdoor scene. 116 IDs are randomly
selected to form the training set, while the remaining 115 IDs are constituted the testing set.
We test the matching rates of this dataset under two training strategies respectively. The results are shown in the TABLE \ref{tab6}.
It can be summarized that with hybrid training, notable improvements can be obtained on both attribute recognition and person Re-ID in our method.


\subsubsection{The contribution of individual attribute}
\begin{figure}[t]
\begin{center}
\includegraphics[width=1\linewidth]{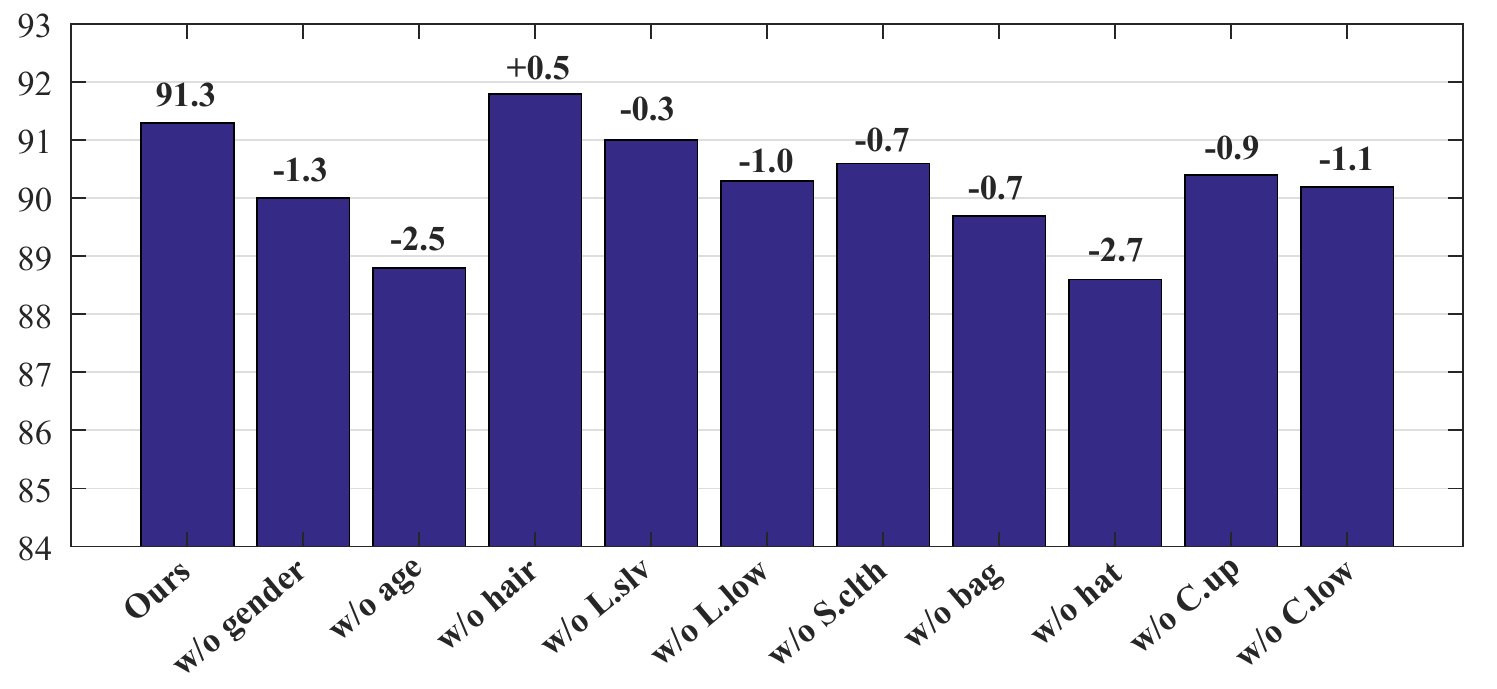}
\end{center}
   \caption{Re-ID Rank-1 accuracy on Market-1501 dataset. One attribute is removed at a time. Notations are the same with
TABLE \ref{tab2}.}
\label{fig:7}
\end{figure}
We evaluate the contribution of individual attributes on the re-ID accuracy. We remove one attribute at a time, and the results on the Market-1501 dataset are summarized in Fig. \ref{fig:7}.
We find that for the 10 attributes on Market-1501,  most of them are indispensable. The most influencing attributes on the datasets are `age' and `hat',
which lead to  Rank-1 decrease of 2.5\% and 2.7\%, respectively. In particular, when we remove an attribute,
the attribute number and order in the mapping table may change greatly, it is shown from the results that it does not affect the performance.
Therefore, we can infer that our method is not influenced by the order of attributes in the attribute mapping table.

\section{Conclusion}

In this paper, a novel attribute processing method, joint CTC-Attention model, is presented to better obtain the relationships among different attributes.
JCM encodes both the images and the attributes into sequences,
and the CTC loss is adopted to improve the encoding performance of the network.
In addition, the attention model that can process sequences in parallel and align sequences is applied using its language learning ability.
Therefore, it can output all predicted attribute values with arbitrary length at a time.
Experiments on three datasets can demonstrate that the proposed method has better performance and higher efficiency than the state of the arts.

{\small
\bibliographystyle{ieee}
\bibliography{egbib}
}

\end{document}